\documentclass[11pt,a4paper]{article}
\usepackage[hyperref]{emnlp-ijcnlp-2019}
\usepackage{times}
\usepackage{graphicx}
\usepackage{url} 
\usepackage{latexsym}
\usepackage{amsmath}
\usepackage{float}
\usepackage{url}
\usepackage{soul}

\aclfinalcopy 

\title{Structuring Latent Spaces for Stylized Response Generation}

\author{Xiang Gao \quad\quad Yizhe Zhang \quad\quad Sungjin Lee \thanks{~~Now at Alexa AI, Amazon.} \\ \quad\quad \textbf{Michel Galley} \quad\quad\textbf{Chris Brockett}\quad\quad \textbf{Jianfeng Gao}\quad\quad \textbf{Bill Dolan}\\
  Microsoft Research, Redmond, WA, USA \\
  {\small \tt \{xiag,billdol\}@microsoft.com}
}

\date{}

\begin{document}
\maketitle
\begin{abstract}
Generating responses in a targeted style is a useful yet challenging task, especially in the absence of parallel data. With limited data, existing methods tend to generate responses that are either less stylized or less context-relevant. We propose \textsc{StyleFusion}, which bridges conversation modeling and non-parallel style transfer by sharing a structured latent space. This structure allows the system to generate stylized relevant responses by sampling in the neighborhood of the conversation model prediction, and continuously control the style level. We demonstrate this method using dialogues from Reddit data and two sets of sentences with distinct styles (arXiv and Sherlock Holmes novels). Automatic and human evaluation show that, without sacrificing appropriateness, the system generates responses of the targeted style and outperforms competitive baselines. \footnote{An implementation of our model and the scripts to generate the datasets are available at \url{https://github.com/golsun/StyleFusion}.}

\end{abstract}

\section{Introduction}
\label{sec:intro}

A social chatbot designed to establish long-term emotional connections with users must generate responses that not only match the \textit{content} of user input and context, but also do so in a desired target \textit{style} \cite{xiaoice, li2016persona, luan2016multiplicative,gao2019neural}.
A conversational agent that speaks in a polite, professional tone is likely to facilitate service in customer relationship scenarios; likewise, an agent that sounds like an cartoon character or a superhero can be more engaging in a theme park. 
The master of response style is also an important step towards human-like chatbots. As highlighted in social psychology studies \cite{niederhoffer2002linguistic, niederhoffer2002sharing}, when two people are talking, they tend to match linguistic style of each other, sometime even regardless of their intentions. 
Achieving this level of performance, however, is challenging.
Lacking parallel data in different conversational styles, researchers often resort to what we will term \textit{style datasets} that are in non-conversational format (e.g. news, novels, blogs). Since the contents and formats of these are quite different from conversation data, existing approaches tend to generate responses that are either less style-specific  \cite{luan2017mtask} or less context-relevant \cite{niu2018polite}.

\begin{figure}[t]
    \centering
    \includegraphics[width=0.47\textwidth]{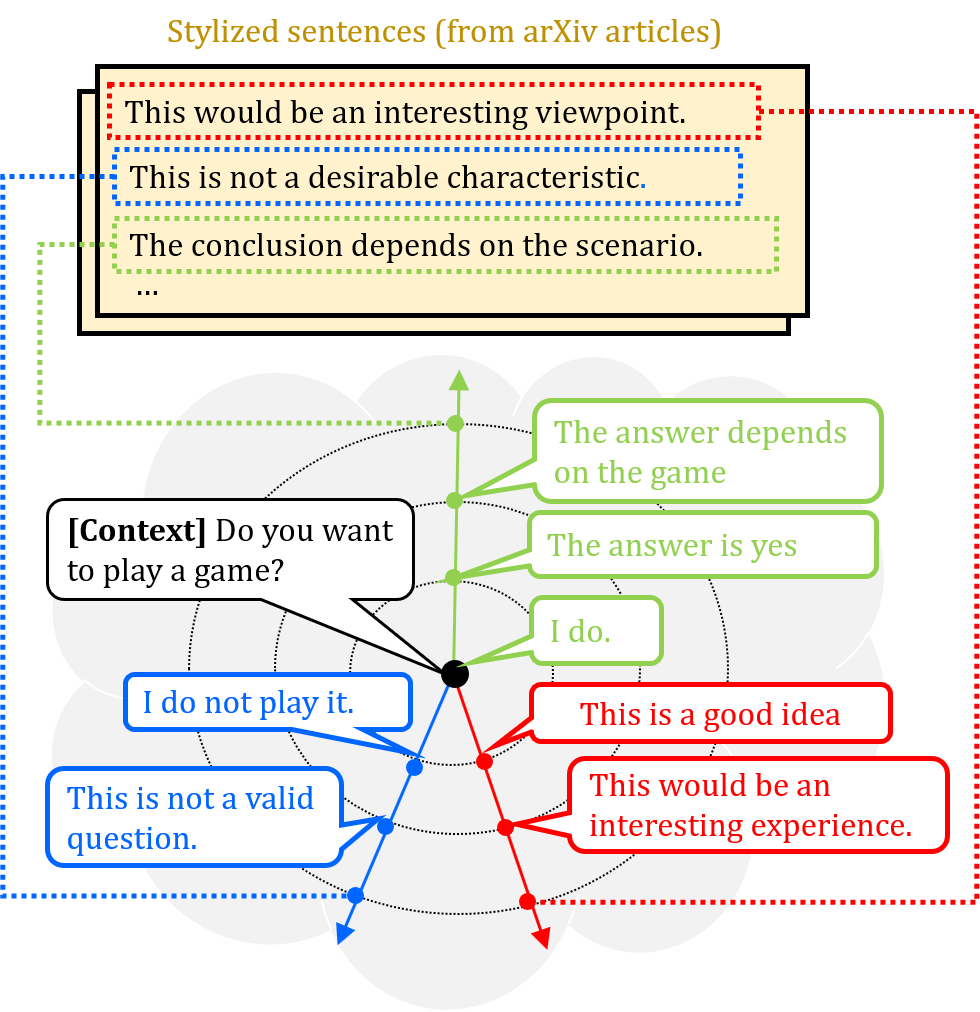}
    \caption{\textsc{StyleFusion} helps conversational model to distill style from non-conversational, non-parallel sentences by mapping them to points surrounding the related conversations in the structured latent space. Direction and distance from the model prediction (center black dot) roughly correspond to contents and style intensity, respectively, illustrated by examples taken from Table~\ref{table:example_towards}.}
    \label{fig:intro}
\end{figure}

We suggest that this trade-off between appropriateness and style stems from profound differences between conversation and style datasets in format, style and contents that impede joint learning. One approach has been to combine these only during decoding:
\citet{niu2018polite} trained two models separately, a Sequence-to-Sequence (S2S) \cite{sutskever2014sequence} on a conversation dataset and a language model (LM) on a style dataset. At inference time, they take a weighted average of the token probability distribution of the two models to predict the next token. This forced bias, however, degrades output relevance.
An alternative approach attempts to map the two datasets into the same latent space:  
\citet{luan2017mtask} use multi-task learning to train a S2S model on a conversation dataset and an autoencoder (AE) on a style dataset. \citet{jointly2019gao} point out that the two datasets still form separate clusters in the latent space; below we observe that this leads to a low style intensity in generated responses (Section~\ref{sec:results}).

We propose to bridge conversation modeling and non-parallel style transfer by structuring a shared latent space using novel regularization techniques, that we dub \textsc{StyleFusion}. In contrast to \citet{luan2017mtask}, the two datasets are well aligned in the latent space by generalizing \textsc{SpaceFusion} \footnote{Integrated into Microsoft Icecaps toolkit \cite{shiv2019microsoft} \url{https://github.com/microsoft/icecaps}. } \cite{jointly2019gao}, an approach that aligns latent spaces for paired samples, to non-parallel datasets. 
In the structured shared latent space, stylized sentences are nudged adjacent to semantically related conversations, thereby allowing the system to generate style-specific relevant responses by sampling in the neighborhood of the model prediction. 
Distance and direction from the model prediction roughly match the style intensity and content diversity of generated responses, respectively, as illustrated in Fig.~\ref{fig:intro}

We demonstrate this method using dialogues from Reddit data and two sets of sentences with distinct styles (arXiv and Sherlock Holmes novels). Automatic and human evaluation show that, without sacrificing appropriateness, our system can generate responses in a targeted style and outperforms competitive baselines.

Our contribution can be summarized thus: 1) We introduce an end-to-end approach that generates style-specific responses from conversational data and non-parallel non-conversation style data. 2) We generalize the \textsc{SpaceFusion} model  of \cite{jointly2019gao} to non-parallel data by a new regularization method. 3) We present a visualization analysis that provides intuitive insights into the drawbacks of alternative approaches.

\section{Related Work}

\paragraph{Text style transfer} is a related but distinct task. It usually preserves the content \cite{yang2018unsupervised,hu2017toward,fu2018style,shen2017style,gong2019reinforcement}. In contrast, content of conversational responses in a given context can be semantically diverse.  
Various approaches have been proposed for non-parallel data setup. \citet{fu2018style} proposed to use separate decoders for different styles and a classifier to measure style strength. \citet{shen2017style} proposed to map texts of two different styles into a shared latent space where the "content" information is preserved and "style" information is discarded. An adversarial discriminator is used to align the latent spaces of two different styles. However, \citet{yang2018unsupervised} point out the difficulty of training an adversarial discriminator and proposed instead the use of language models as discriminator. 
Like \citet{shen2017style, yang2018unsupervised}, we align latent spaces for different styles. However we also align latent spaces encoded by different models (S2S and AE). 

\paragraph{Stylized response generation} is a relatively new task. \citet{akama2017stylecons} use a stylized conversation corpus to fine-tune a conversation model pretrained on a background conversation dataset. However, stylized texts are usually in non-conversational format, as in the present setting. \citet{niu2018polite} proposed a method that takes the weighted average of the token probability distribution predicted by a S2S trained on background conversational dataset and that predicted by a LM trained on style dataset as the token probability. They observed reduced relevance and attributed this to the fact that the LM was not trained to attend to conversation context and S2S was not trained to learn style during training. 
In contrast, we jointly learn from conversation and style datasets during training.
\citet{niu2018polite} have proposed label-fine-tuning, but this is limited to scenarios where a reasonable portion of the conversational dataset is in the target style, which is not always the case.

\paragraph{Persona-grounded conversation modeling} \citet{li2016persona, luan2017mtask} aim to generate responses mimicking a speaker. It is closely related to the present task, since persona is, broadly speaking, the manifestation of a type of style. \citet{li2016persona} feeds a speaker ID to the decoder to promote generation of response for that target speaker. However non-conversational data cannot be used. \citet{luan2017mtask} applied a multi-task learning approach to utilize non-conversational data. A S2S model, taking in conversational data, and an autoencoder (AE), taking in non-conversational data, share the decoder and are trained alternately. However, \citet{jointly2019gao} observed that sharing the decoder may not truly allow S2S and AE to share the latent space, and thus S2S may not fully utilize what is learned by AE. Unlike \citet{li2016persona} using labelled persona IDs, \citet{yizhe2019consistent} have proposed using a self-supervised method to extract persona features from conversation history. This allows modeling persona dynamically, which agrees with the fact that even the same person can speak in different style in different scenarios. 

\paragraph{Multi-task learning} \citet{McCann2018decaNLP, mt-dnn, luan2017mtask, jointly2019gao, zhang2017deconvolutional} aggregates the strengths of each specific task, and induces regularization effects \cite{mt-dnn} as the model is trained to learn a more universal representation. However a simple multi-task approach \cite{luan2017mtask} may learn separate representations for each dataset \cite{jointly2019gao}. To address this, in previous work \cite{jointly2019gao}, we proposed the \textsc{SpaceFusion} model featuring a regularization technique that explicitly encourages alignment of latent spaces for a universal representation. \textsc{SpaceFusion}, however, is only designed for paired samples. We generalize \textsc{SpaceFusion} to non-parallel datasets in this paper.

\section{The \textsc{StyleFusion} Model}
\subsection{Problem statement}
\label{sec:def}
Let $\mathcal{D}_{conv}=[(x_0,y_0),(x_1,y_1),\cdots,(x_n,y_n)]$ denote a conversation dataset, where $x_i$ and $y_i$ are context sentences and a corresponding response, respectively. $x_i$ consists of one or more utterances and $y_i$ is only one utterance. 
$\mathcal{D}_{style}=[s_0,s_1,\cdots,s_m]$ denotes a non-conversational style dataset, where $s_i$ is a sentence sampled from a corpus of the targeted style. Samples from $\mathcal{D}_{style}$ do not have a labelled corresponding relation with samples from $\mathcal{D}_{conv}$ (thus "non-parallel"). 
Our aim is to train a model jointly on $\mathcal{D}_{style}$ and $\mathcal{D}_{conv}$ to generate appropriate responses in the style similar to sentences from $\mathcal{D}_{style}$, to a given context. The iven context may or may not be in the target style.

\subsection{Training}

In contrast to \textsc{SpaceFusion}\cite{jointly2019gao}, which only fuses context-response pairs, our goal is to additionally map related stylized sentences to points surrounding the context in the shared latent representation space. The system can then generate relevant stylized responses by sampling in the neighborhood of the prediction based on the context. 

\begin{figure}[tb]
    \centering
    \includegraphics[width=0.47\textwidth]{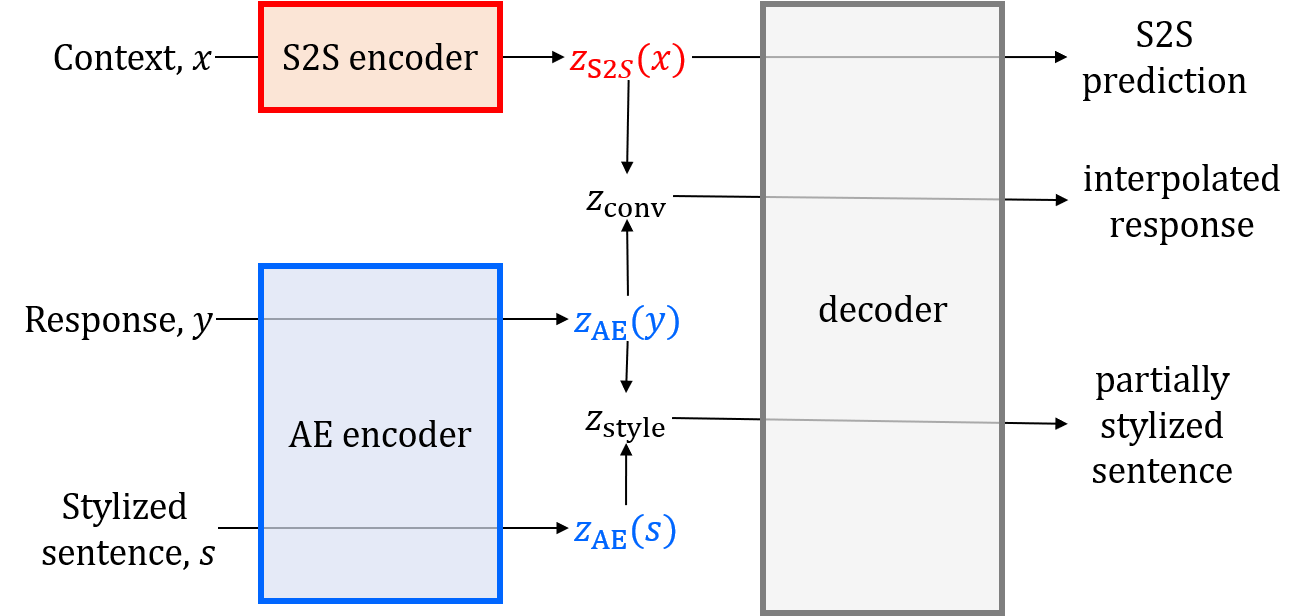}
    \caption{\textsc{StyleFusion} model architecture.}
    \label{fig:model_train}
\end{figure}

\paragraph{Model overview.}
As illustrated in figure~\ref{fig:model_train}, the model consists of a sequence-to-sequence (S2S) module and an autoencoder (AE) module that shares a decoder. We use S2S encoder to produce the prediction representation $z_{S2S}(x_i)$, or "latent action" \cite{tony2019thesis}, and AE encoder to obtain the representation of the corresponding responses $z_{AE}(y_i)$ and stylized sentences $z_{AE}(s_j)$. We use generalized regularization terms, fusion and smoothness, to align the three latent spaces  $z_{S2S}(x_i)$, $z_{AE}(y_i)$, and $z_{AE}(s_j)$.

\paragraph{Fusion objectives} encourage different latent spaces to be close to each other. Accordingly, we define the cross-latent-space distances to be minimized. 
For response appropriateness, as $x_i$ and $y_i$ are paired as context and response, we use their pair-wise dissimilarity, following \citet{jointly2019gao}
\begin{align}  
\label{eq:dist_conv}  
    d_{\text{conv}} = \sum_{i\in \text{batch}} \frac{d_E( z_{\text{S2S}}(x_i),z_{\text{AE}}(y_i))}{n\sqrt{l}}
\end{align}
where $n$ is the batch size, $l$ is the dimension of latent space 
, and we use $d_E$, the Euclidean distance in latent space, as the dissimilarity.

For style transfer, however, $x_i$ and $s_j$ is not paired. Thus, we instead minimize the distance between a point and its nearest neighbor from another dataset to pull these two datasets close to each other in the shared latent space. 
\begin{align}  
\label{eq:d_style}  
    d_{\text{style}} =& \frac{1}{2} d_{\text{NN}}^{\text{ cross}}(\{z_{\text{S2S}}(x_i)\}, \{z_{\text{AE}}(s_i)\}) + \nonumber \\
    & \frac{1}{2} d_{\text{NN}}^{\text{ cross}}(\{z_{\text{AE}}(s_i)\}, \{z_{\text{S2S}}(x_i)\})
\end{align}
where $d_{NN}^{\text{ cross}}(\{a_i\}, \{b_i\})$ is the batch average of the distance between $a_i$ and $b_{\text{NN of }a_i}$ -- the nearest neighbor (NN) of $a_i$ from set $\{b_i\}$
\begin{align}  
\label{eq:d_NN}  
    d_{\text{NN}}^{\text{ cross}}(\{a_i\},\{b_i\}) =&  \sum_{i\in \text{batch}}  \frac{d_E(a_i,b_{\text{NN of }a_i})}{n\sqrt{l}}
\end{align}
While minimizing the cross-latent-space distances, $d_{\text{conv}}$ and $d_{\text{style}}$, we want the samples from the same latent space spread out, following \citet{jointly2019gao}. For this purpose, \citet{jointly2019gao} maximized the average of capped distance between points from the same latent space. However, we found that the results are sensitive to the cap value. Instead, we define the following nearest-neighbor-based characteristic distance
\begin{align}  
\label{eq:d_conv}  
    d_{\text{spread-out}} =&  \min[d_{\text{NN}}^{\text{ same}}(\{z_{\text{AE}}(y_i)\}), \nonumber\\
    & d_{\text{NN}}^{\text{ same}}(\{z_{\text{AE}}(s_i)\}), \nonumber\\
    & d_{\text{NN}}^{\text{ same}}(\{z_{\text{S2S}}(x_i)\})] \\
    d_{\text{NN}}^{\text{ same}}(\{a_i\}) =& \sum_{i\in \text{batch}} \frac{d_E(a_i,a_{\text{NN of }a_i})}{n\sqrt{l}}
\end{align}

Combining these loss terms we have the following two objectives:
\begin{align}  
\label{eq:dist}  
    \mathcal{L}_{\text{fuse,conv}} =& d_{\text{conv}} - d_{\text{spread-out}} \\
    \mathcal{L}_{\text{fuse,style}} =& d_{\text{style}} - d_{\text{spread-out}}
\end{align}

\paragraph{Smoothness objective} encourages smooth semantic transition in the shared latent space. For response appropriateness, following \citet{jointly2019gao}, we encourage the interpolation between the prediction $z_{S2S}(x_i)$ and the target response $z_{AE}(y_i)$ to generate the target response $y_i$. 
\begin{align}  
\label{eq:smooth_conv}  
    z_{\text{conv}} = (1-u)z_{\text{AE}}(y) + u z_{\text{S2S}}(x) + \epsilon \\
    \mathcal{L}_{\text{smooth,conv}} = - \frac{1}{|y|} \log p(y|z_{\text{conv}})
\end{align}
where $u \sim U(0,1)$ is a uniformly distributed random variable, and $ \epsilon$ is a Gaussian noise with zero mean and covariance matrix of $\sigma^2I$.

For style transfer, as we move from a non-stylized sentence $z_{\text{AE}}(x)$ to a random stylized sentence $z_{\text{AE}}(s)$, we expect to generate a partially stylized sentence and encourage the generated sentence to gradually change from $x$ to $s$.
\begin{align}  
    z_{\text{style}} = & (1-u)z_{\text{AE}}(x) + u z_{\text{AE}}(s) + \epsilon
\end{align} 
\begin{align}  
\label{eq:smooth_style}  
    \mathcal{L}_{\text{smooth,style}} = & - (1-u) \frac{1}{|x|} \log p(x|z_{\text{style}}) \nonumber\\
    & - u \frac{1}{|s|} \log p(s|z_{\text{style}})
\end{align} 
\paragraph{Training objective} to be minimized is a combination of a vanilla S2S and the above regularization terms \footnote{More generally, one may use a weighted sum of these terms instead. We set them equally weighted for simplicity}. $\mathcal{L}_{\text{smooth,style}}$ and $\mathcal{L}_{\text{fuse,style}}$ are new terms not existing in \cite{jointly2019gao}. A more compact definition $\mathcal{L}_{\text{conv}} = \mathcal{L}_{\text{smooth,conv}} + \mathcal{L}_{\text{fuse,conv}}$ and $\mathcal{L}_{\text{style}} = \mathcal{L}_{\text{smooth,style}} + \mathcal{L}_{\text{fuse,style}}$ yields
\begin{align}  
\label{eq:final_loss}  
    \mathcal{L} =   - \frac{1}{|y|} \log p(y|z_{\text{S2S}}) + \mathcal{L}_{\text{conv}} + \mathcal{L}_{\text{style}}
\end{align}
For the case $\mathcal{D}_{style}$ is much smaller than $\mathcal{D}_{conv}$, as in the present work, the model may overfit on $\mathcal{D}_{style}$. We propose to firstly pretrain the model on $\mathcal{D}_{conv}$ only \footnote{by setting terms $\mathcal{L}_{\text{smooth,style}}$ and $\mathcal{L}_{\text{fuse,style}}$ as zero}, then continue training on both $\mathcal{D}_{conv}$ and $\mathcal{D}_{style}$. Furthermore, to reduce overfitting, we applied a data augmentation technique by randomly mask tokens in $s_i$ by a special out-of-vocab token. The masking probability of a token is inversely proportional to its frequency in training data. 

\subsection{Inference}
\label{sec:infer}
Following \cite{jointly2019gao}, we sample in the neighborhood of $z_{S2S}(x)$ by adding a noise $r$ of a given length $|r|$ towards a direction randomly drawn from the uniform distribution. 
\begin{align}  
\label{eq:z+r}  
    z = z_{\text{S2S}}(x) + r
\end{align}
As $r$ depends on $l$, the dimension of $z$, We define a normalized value
\begin{align}  
\label{eq:rho}  
    \rho = |r|/(\sigma \sqrt{l})
\end{align}
As the stylized texts are usually sparse, it is possible to generate non-stylized hypothesis as we vary $\rho$ along some direction. Thus we rank the hypotheses considering both relevance and style intensity.
\begin{align}  
\label{eq:rank}  
    \text{score}(h_i) = (1-\lambda) P(h_i|z_{\text{S2S}}(x)) + \lambda P_{\text{style}}(h_i)
\end{align}
where $\lambda$ = 0.5 unless otherwise specified, $P(h_i|z_{\text{S2S}}(x))$ estimates the relevancy, and $P_{style}(h_i)$ is the probability of hypothesis $h_i$ being targeted style predicted by pretrained classifiers. 

We considered two style classifiers: a "\textbf{neural}" based on two stacked GRU \cite{cho2014gru} cells, and a "\textbf{ngram}" classifier which is a logistic regressor using ngram (n=1,2,3,4) multi-hot features. Both classifiers are trained using $\{y_i\}$ as negative samples and $\{s_i\}$ as positive samples. 
$P_{style}(h_i)$ is computed by taking average of the prediction of these two classifiers.

\section{Experiment Setup}

\subsection{Tasks and datasets}
\label{sec:datasets}
We experiments with two tasks: generating arXiv-like and Holmes-like responses, respectively, using the datasets summarized in Table~\ref{table:datasets}

\begin{table}[h]
    \centering
    \small
    \begin{tabular}{p{0.1\textwidth}|p{0.05\textwidth}p{0.06\textwidth}|p{0.17\textwidth}}
    \hline
    Task & \multicolumn{2}{l|}{Training} & Testing \\
          & $\mathcal{D}_{\text{conv}}$ & $\mathcal{D}_{\text{style}}$ & $\mathcal{D}_{\text{test}}$ \\
    \hline
    arXiv-like & Reddit & arXiv & arXiv-like Reddit \\
    Holmes-like & Reddit & Holmes & Holmes-like Reddit \\
    \hline
    \end{tabular}
    \caption{Summary of tasks and datasets}
    \label{table:datasets}
\end{table}

(i) \textit{Reddit} is a conversation dataset constructed from posts and comments on Reddit.com \footnote{using raw data collected by a third party \url{http://files.pushshift.io/reddit/comments/}} during 2011, consisting of 10M pairs of context and response .
(ii) \textit{arXiv} is a non-conversational dataset extracted from articles on arXiv.org \footnote{from KDD 2003 \url{http://www.cs.cornell.edu/projects/kddcup/datasets.html}} from 1998 to 2002, consisting of 1M sentences .
(iii) \textit{Holmes} refers to another non-conversational dataset extracted from Sherlock Holmes novel series\footnote{from \url{https://gutenberg.org}}
by Arthur Conan Doyle, with 38k sentences.

$\mathcal{D}_{\text{test}}$ is the test set with stylized reference responses, constructed by filtering the Reddit dataset from year 2013 using the trained neural and ngram classifiers. For each context, there are at least 4 reference responses approximately in the targeted style ($P_{\text{style}} > 0.3$). The style intensity of the context is not filtered.

\subsection{Human evaluation}

We designed the following two tasks.
\paragraph{Response appropriateness} measurement task presents a context and a set of hypotheses (from the present and baseline systems), and for each hypothesis annotators choose from one of the following options that best fits the quality of the response: ok, marginal, bad (generic or irrelevant), and then map them to numerical score 1, 0.5, and 0, respectively.
\paragraph{Style classification} task presents a hypothesis and two groups of example sentences, from Reddit and style corpus (Holmes or arXiv). Then crowd-sourced annotators judge whether the hypothesis is more similar to the Reddit group, not sure, or more similar to the style corpus group. We then map these to numerical scores 0, 0.5, and 1, respectively.

For all tasks, the hypotheses of different systems of the same set of 500 randomly selected $x_i$ are presented in random order and the identity of the system is invisible to annotators. Each sample is judged by 5 annotators individually.

\subsection{Automatic evaluation}
\label{sec:metrics}
We measure relevance using multi-reference BLEU \citet{papineni2002bleu}, and diversity by entropy 4-gram \cite{zhang2018gan}, and distinct 1,2-gram \cite{li2016mmi}. 

For style intensity evaluation, besides the \textbf{neural} and \textbf{ngram} classifier prediction (Section~\ref{sec:infer}), we also use simple word-counting (hereafter \textbf{count} metric) to minimize model-specific effects. We first construct a training corpus with balanced positive (from $\mathcal{D}_{\text{style}}$) and negative (responses sampled from $\mathcal{D}_{\text{conv}}$) samples. Then, for each word that appears in more than 5 sentences in the training corpus, we compute the average style intensity of sentences that contain this word. The top $k$ words of highest style intensity are chosen as the keywords in this style. For a test corpus, we compute the average ratio of words that are keywords of a style, as its "count" style metric. 

Besides the overall style comparisons (Reddit vs. Holmes, and Reddit vs. arXiv), we also crowd-sourced three sets of sentences with human labeled levels in three finer styles: how formal, emotional, and technical each sentence is, and build the keyword list for the count metric.

\subsection{Baselines}
We compare the following baseline systems.

The first category is generative models. 
(i) \textit{MTask} refers to the vanilla multi-task learning model proposed in \cite{luan2017mtask} trained on both $\mathcal{D}_{\text{conv}}$ and $\mathcal{D}_{\text{style}}$. 
(ii) \textit{S2S+LM} refers to the method proposed by \citet{niu2018polite}\footnote{This method was referred as "Fusion" in \cite{niu2018polite} but to avoid confusing readers with our \textsc{StyleFusion} method, we refer it as "S2S+LM"}, which uses the weighted average of a S2S model, trained on $\mathcal{D}_{\text{conv}}$, and a LM model, trained on $\mathcal{D}_{\text{style}}$, as the token probability distribution at inference time.

The second category draws a training sample as hypothesis. 
(iii) \textit{Retrieval} refers to a simple retrieval system which returns the sentence from $\mathcal{D}_{\text{style}}$ of highest generation probability by the MTask model.
(iv) \textit{Rand} is a system that randomly picks a sentence from $\mathcal{D}_{\text{style}}$. 
(v) \textit{Human} refers to the system randomly picks one of the multiple reference responses in the given context from $\mathcal{D}_{\text{test}}$.

\subsection{Model setup}
\textsc{StyleFusion} and trainable baselines, MTask and S2S+LM, use two stacked GRU \cite{cho2014gru} cells for encoders and decoders with $l=1000$. The word embedding is also 1000 dimension, trained from random initialization. The variance of the noise $\epsilon$ is set to $\sigma^2=0.1^2$. 
The state of the top layer of encoder GRU cell is used as $z$. $z$ is used as the initial state of all layers of the decoder.
All trainable models are trained with the ADAM method \cite{kingma2014adam} with a learning rate of 0.0003. For \textsc{StyleFusion} and MTask, we first train on $\mathcal{D}_{\text{conv}}$ for 2 epochs, and then continue the training on both $\mathcal{D}_{\text{conv}}$ and $\mathcal{D}_{\text{style}}$ until convergence \footnote{approximately one pass of arXiv and 5 passes of Holmes}. For all systems except "Rand" and "Retrieval", we use the ranking method Eq~\ref{eq:rank} to select the top one hypothesis from 100 candidates.

\section{Results and Analysis}
\label{sec:results}

\begin{table}[ht]
    \centering
    \small
    \begin{tabular}{p{0.1\textwidth}|p{0.33\textwidth}} \hline 
    \textbf{context} & Do you want to play a game?\\
    \hline
    \textbf{towards} & The conclusion depends on the scenario .\\
    $\rho=0.0$ & I do. \\
    $\rho=0.5$ & The answer is yes. \\
    $\rho=1.0$ & The answer depends on the game. \\
    \hline
    \textbf{towards} & This would be an interesting viewpoint.\\
    $\rho=0.4$ & This is a good idea. \\
    $\rho=0.9$ & This would be an interesting experience\\
    \hline
    \textbf{towards} & This is not a desirable characteristic.\\
    $\rho=0.5$ & I don't play it. \\
    $\rho=1.0$ & This is not a valid question.\\
    \hline
    \end{tabular}
    \caption{Example \textsc{StyleFusion} outputs for arXiv-like response generation task at different distance $\rho$ and direction (towards $z_\text{AE}$ of a given sentence)}
    \label{table:example_towards}
\end{table}

\subsection{Modulating the style}

By leveraging the structure of the shared latent space, we can modulate the style intensity by $\rho$, as illustrated by examples in Table~\ref{table:example_towards}. For example, given the context \textit{"Do you want to play a game"}, the hypothesis generated from $\rho=0$ is \textit{"I do"}, which is non-stylized. While moving towards $z_\text{AE}$ of an arXiv-style sentence \textit{"This would be an interesting viewpoint"}, the responses gradually change to "This would be an interesting experience" at $\rho=1.0$, which remains relevant but is more similar to the target style. Similar trends can be observed when moving in the other direction \textit{"The conclusion depends on the scenario"} and \textit{"This is not a desirable characteristic"}. It also shows that the contents are affected by the direction, a desired property inherited from \textsc{SpaceFusion} models. 

\begin{figure}[tb]
    \centering
    \includegraphics[width=0.47\textwidth]{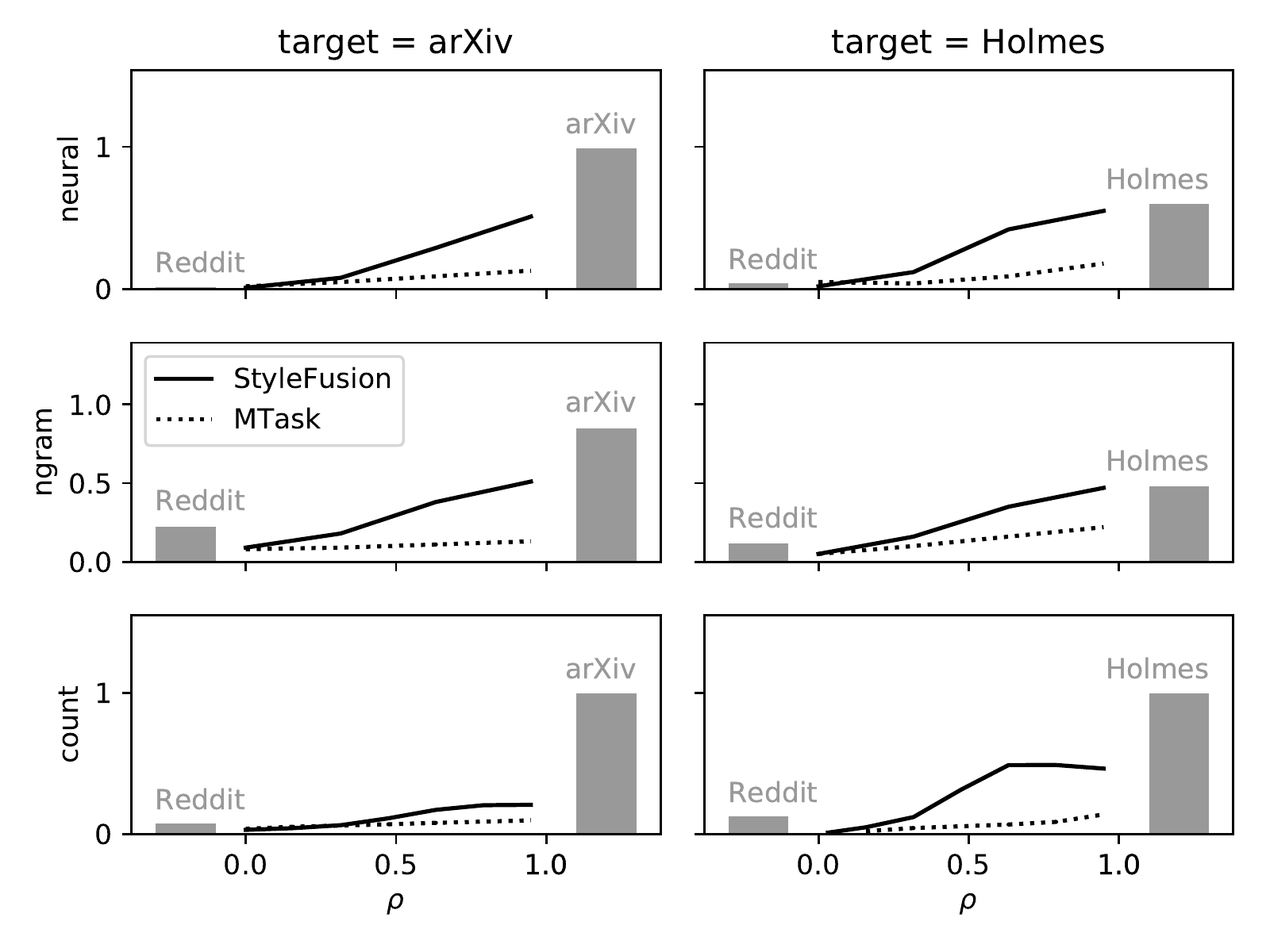}
    \caption{Change of the overall style intensity with $\rho$, as measured by two classifiers "neural" and "ngram" (Section~\ref{sec:infer}) and the "count" metric. The "count" metric is normalized by the value of the target style corpus. The barplot shows the desired trend (from Reddit to arXiv or Holmes), and the lines the actual trends.}
    \label{fig:overall_vs_rho}
\end{figure}

The relation between style intensity and $\rho$ is further confirmed by automatic measurement. As illustrated in Fig.~\ref{fig:overall_vs_rho}, as $\rho$ increases, responses come to resemble the targeted style within the depicted range. In contrast, the style intensity of MTask outputs rises only slightly as $\rho$ increases. 

\begin{figure}[tb]
    \centering
    \includegraphics[width=0.47\textwidth]{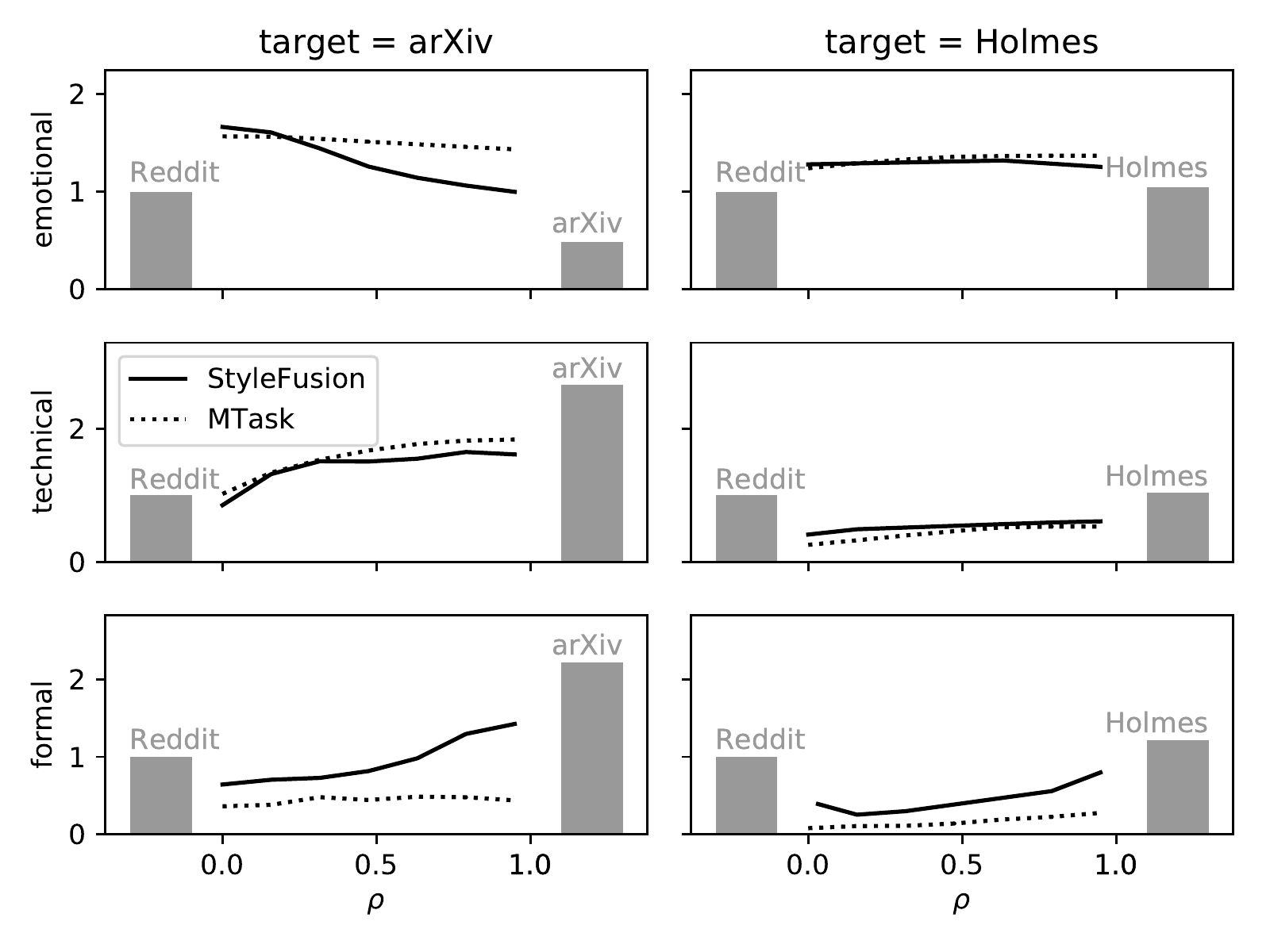}
    \caption{Change of the styles at finer granularity of $\rho$, measured by the "count" metric normalized by the value of Reddit dataset. The barplot shows the desired trend (from Reddit to arXiv or Holmes), and the lines the actual trends.}
    \label{fig:finer_vs_rho}
\end{figure}

The increase of overall style intensity is coupled with change in the style's finer granularity, as illustrated in Fig.~\ref{fig:finer_vs_rho}. Compared to Reddit, arXiv is less emotional, and more formal and technical. Consistent with this, \textsc{StyleFusion} outputs exhibit less emotion, but become much more technical and formal as $\rho$ increases. MTask, however, tends only to show increased technical level, but fails to be less emotional and more formal, inconsistent with the target style. Where Holmes is the target, the emotional and technical levels do not significantly change compared to Reddit, but Holmes is stylistically more formal. \textsc{StyleFusion} captures these trends, whereas MTask outputs are insufficiently formal, shown in Fig.~\ref{fig:finer_vs_rho}(lowest panel).

\begin{figure}[tb]
    \centering
    \includegraphics[width=0.47\textwidth]{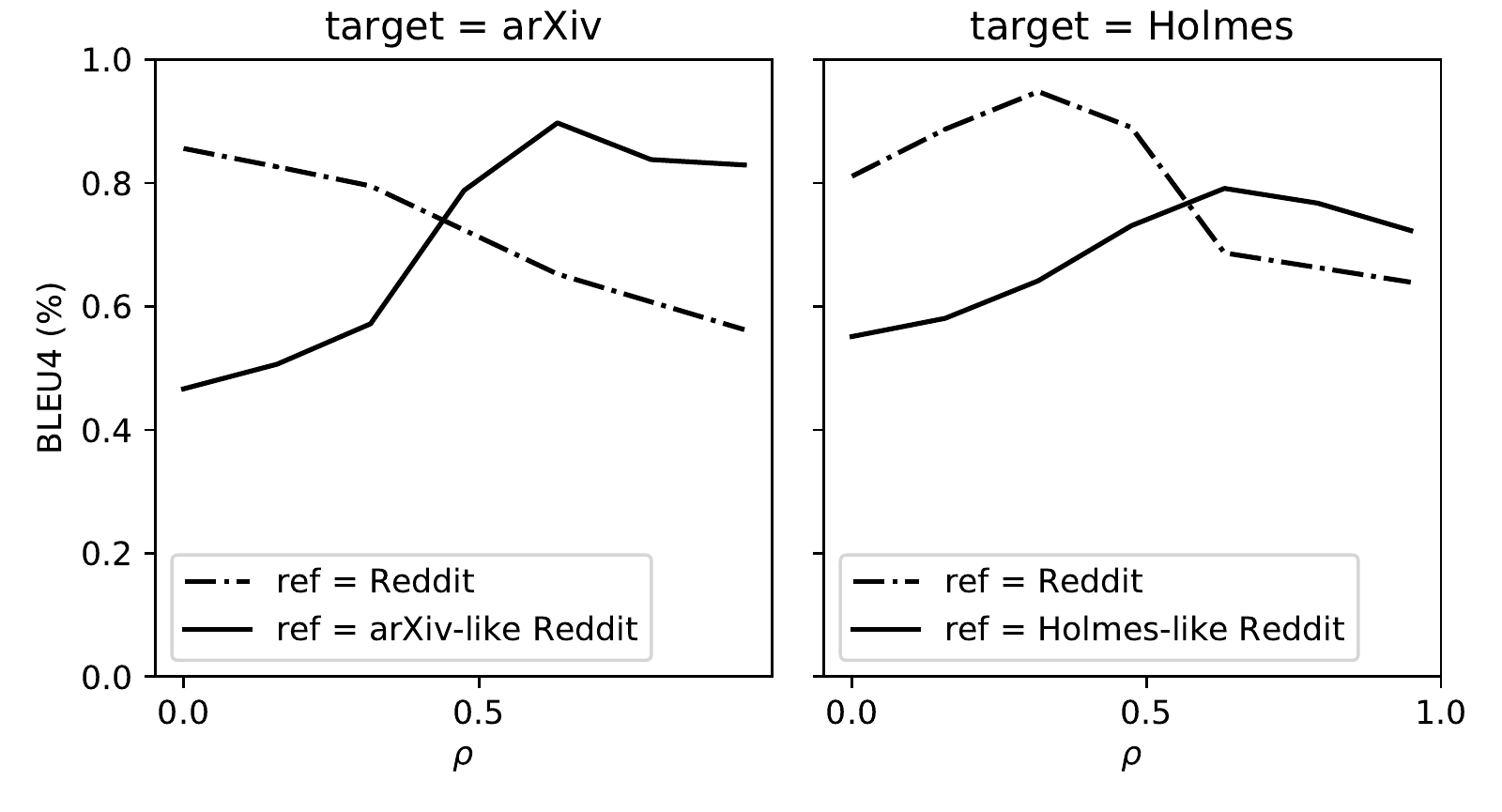}
    \caption{Relevancy of the \textsc{StyleFusion} outputs at different $\rho$ as measured by BLEU4 with references of different styles}
    \label{fig:bleu_vs_rho}
\end{figure}

We also measured the BLEU4 score at different $\rho$ as shown in Fig.~\ref{fig:bleu_vs_rho}. Besides $\mathcal{D}_{\text{test}}$ (stylized references), we also tested on held-out Reddit data (i.e. non-stylized references). We observed that at smaller $\rho$, the \textsc{StyleFusion} outputs are relevant to context and style is more similar to Reddit, indicated by the relatively high BLEU4 computed using non-stylized references. At greater radius, BLEU4 rises when calculated on arXiv-like and Holmes-like references, indicating that the responses generated remain relevant but are closer to the targeted style. Combined with the case at small radius, the implication is that, although the style gradually changes, the responses generated by \textsc{StyleFusion} remain relevant at a relatively wide range of radius.

\begin{table}[ht]
    \centering
    \small
    \begin{tabular}{p{0.06\textwidth}p{0.37\textwidth}} \hline 
    \textbf{context} & Okay, but can we host it in the cloud?\\
    \hline
    \textsc{Style-Fusion}  & It would be an interesting experiment. It is a possibility. \\
    S2S &   I think it might be a bit of a stretch.\\
    MTask & Yes, yes you can.\\
    S2S+LM &  The relevant question is , as to this , is you referring to the casimir energy program?\\
    Retrieval &   Even though this issue can be analyzed in more detail, we shall content ourselves with some simple statements. \\
    rand &   In particular, it involves the combination equation\\
    human & To the cloud !\\
    \hline
    \textbf{context} & The universe has no edge.\\
    \hline
    \textsc{Style-Fusion}  & Does this mean the universe is expanding? \\
    S2S &  The universe is infinite. \\
    MTask &  The universe is finite.\\
    S2S+LM &  The universe can be on the infinite loop. \\
    Retrieval &  Is there really any difference in the low energy physics? \\
    rand &  In other words , the wave function in equation. \\
    human & So... a sphere? \\
    \hline
    \end{tabular}
    \caption{Example system outputs on arXiv-like response generation task at $\rho=1$}
    \label{table:example_arXiv_baselines}
\end{table}

\begin{table}[ht]
    \centering
    \small
    \begin{tabular}{p{0.06\textwidth}p{0.37\textwidth}} \hline 
    \textbf{context} & First step: go check behind the cabinet and see what's back there.\\
    \hline
    \textsc{Style-Fusion}  & I would advise against this. \\
    S2S &  I'm going to try this next time.\\
    MTask & I'll be back in a few minutes.\\
    S2S+LM & I've got a little time that i had not been engaged yet.\\
    Retrieval &    i was thinking the same thing. \\
    Rand &   Holmes shrugged his shoulders.\\
    human &  Then there would be nothing , and that would be the answer. \\
    \hline
    \textbf{context} & that 's a surprisingly large amount of police coming to see a *** teenager.\\
    \hline
    \textsc{Style-Fusion}  & What is this crime you speak of?\\
    S2S &   I'm pretty sure that 's the only thing that 's going to happen. \\
    MTask &  I'm not sure what you mean by that.\\
    S2S+LM &  I should be on the same business. \\
    Retrieval &   well, yes. \\
    Rand &  I shall be back in an hour or two.   \\
    human & Must have feared what he was packin' \\
    \hline
    \end{tabular}
    \caption{Example system outputs on Holmes-like response generation task at $\rho=1$}
    \label{table:example_holmes_baselines}
\end{table}

\subsection{Fused latent space}

\begin{figure}[tb]
    \centering
    \includegraphics[width=0.47\textwidth]{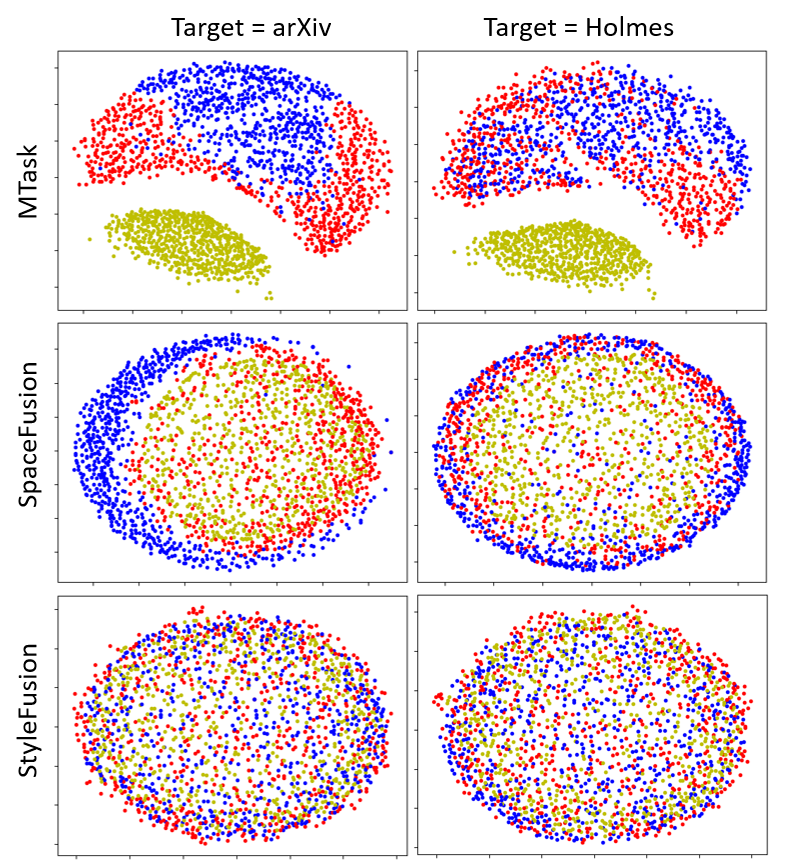}
    \caption{MDS visualization of learned latent spaces. \textbf{\textcolor{yellow}{yellow}} dots for $z_{\text{S2S}}(x)$, \textbf{\textcolor{blue}{blue}} dots for $z_{\text{AE}}(s)$ and \textbf{\textcolor{red}{red}} dots for $z_{\text{AE}}(y)$}
    \label{fig:mds}
\end{figure}

As illustrated by the MDS \cite{borg2003mds} visualization in Fig.~\ref{fig:mds}, MTask cannot align different latent spaces, not only those from different model ($z_{\text{AE}}(y)$ and $z_{\text{S2S}}(x)$), but also for those from same model that have different styles ($z_{\text{AE}}(y)$ and $z_{\text{AE}}(s)$). \textsc{SpaceFusion} \cite{jointly2019gao} can align $z_{\text{AE}}(y)$ and $z_{\text{S2S}}(x)$ better, but $z_{\text{AE}}(s)$ forms a separate cluster, indicating that the conversation dataset and style dataset remain unaligned in the latent space. This is because \textsc{SpaceFusion} was not designed to align non-parallel samples. 
The separation between the conversation dataset and style dataset in latent space, as is the case for MTask and \textsc{SpaceFusion}, makes it difficult for the conversation model to use style knowledge. In contrast, \textsc{StyleFusion} aligns all three latent spaces well as evidenced by Fig.~\ref{fig:mds}.

\subsection{Human evaluation}

\begin{table}[ht]
    \centering
    \small
    \begin{tabular}{p{0.07\textwidth}|p{0.1\textwidth}p{0.06\textwidth}p{0.07\textwidth}p{0.06\textwidth}}
    \hline
    target & model & appropr-iateness & style\textcolor{white}{-}intensity & harmonic mean\\
    \hline
    arXiv & \textsc{StyleFusion} & \textbf{0.17} & 0.26 & \textbf{0.20}\\
          & MTask & \textbf{0.17} & 0.11 & 0.14 \\
          & S2S+LM & 0.09 & 0.51 & 0.15 \\
          & Retrieval & 0.07 & 0.71 & 0.14 \\
          & Rand & 0.04 & \textbf{0.96} & 0.07 \\
          \cline{2-5}
          & Human & 0.51 & 0.28 & 0.36 \\
    \hline
    Holmes & \textsc{StyleFusion} & \textbf{0.22} & 0.28 & \textbf{0.25} \\
          & MTask & \textbf{0.23} & 0.15 & 0.18\\
          & S2S+LM & 0.03 & 0.55 & 0.05 \\
          & Retrieval & 0.14 & 0.30 & 0.19 \\
          & Rand & 0.08 & \textbf{0.69} & 0.14 \\
          \cline{2-5}
          & Human & 0.63 & 0.26 & 0.37 \\
    \hline
    \end{tabular}
    \caption{Human evaluation results. The top models (and those models that are not statistically different, except "human") are in \textbf{bold}.}
    \label{table:human_eval}
\end{table}

Human evaluation results are presented in Table~\ref{table:human_eval}. As in the automatic evaluation results, \textsc{StyleFusion} and MTask show the highest appropriateness (not statistically different) apart from the Human system. However \textsc{StyleFusion} outputs are much more stylized. Rand, Retrieval and S2S+LM tend to generate stylized but irrelevant responses. To make the overall trends sharper, following \cite{jointly2019gao}, we compute the harmonic mean of appropriateness and style intensity, in terms of which \textsc{StyleFusion} outperforms all baselines except the Human system. Additional examples of the system outputs and human responses are provided in Table~\ref{table:example_arXiv_baselines} and Table~\ref{table:example_holmes_baselines}

\begin{table*}[ht]
    \centering
    \small
    \begin{tabular}{p{0.22\textwidth}|p{0.05\textwidth}p{0.05\textwidth}p{0.05\textwidth}|p{0.05\textwidth}p{0.05\textwidth}p{0.05\textwidth}p{0.05\textwidth}|p{0.05\textwidth}p{0.05\textwidth}p{0.05\textwidth}} \hline 
     
    \multicolumn{1}{c|}{system} & \multicolumn{3}{c|}{style intensity}   & \multicolumn{4}{c|}{relevancy}   & \multicolumn{3}{c}{diversity} \\
    sampled ($\in$) or generated  & neural & ngram & count & BLEU1 & BLEU2 & BLEU3 & BLEU4 & entropy4 & distinct1 & distinc2 \\ 
    \hline
    \hline
    \multicolumn{11}{c}{target = arXiv}  \\
    \hline
    Rand ($\in\mathcal{D}_{\text{style}}$)       & {0.99} & {0.85} & {1.00} & 12.1 & 1.7 & 0.6 & 0.34 & {9.4} & {0.13} & {0.56} \\
    Retrieval ($\in\mathcal{D}_{\text{style}}$) & 0.84 & 0.77 & 0.59 & 15.5 & 2.3 & 0.8 & 0.49 & 7.6 & 0.06 & 0.19 \\
    Human ($\in\mathcal{D}_{\text{test}}$)   & 0.43 & 0.47 & 0.35   & 29.0 & 16.3 & 10.6 & 7.44 & 8.6 & 0.31 & 0.81 \\
    \hline
    S2S+LM   & 0.36 & \textbf{0.48} & \textbf{0.34} & 14.2 & 2.5 & 0.7 & 0.41 & \textbf{9.4} & \textbf{0.11} & \textbf{0.54} \\
    MTask   & 0.13 & 0.13 & 0.10 & 16.5 & 2.9 & 1.2 & 0.66 & 5.7 & 0.04 & 0.13 \\
    \multicolumn{1}{r|}{+$\mathcal{L}_\text{conv}$ (\textsc{SpaceFusion})}      & 0.27 & 0.41 & 0.17 & \textbf{18.1} & {3.9} & {1.4} & {0.75} & 6.9 & 0.04 & 0.13 \\ 
    \multicolumn{1}{r|}{+$\mathcal{L}_\text{style}$ (\textsc{StyleFusion})}  & \textbf{0.40} & 0.46 & 0.21 & 17.9 & \textbf{4.4} & \textbf{1.6} & \textbf{0.83} & 7.9 & 0.05 & 0.20 \\ 
    \hline
    \hline
    \multicolumn{11}{c}{target = Holmes}  \\
    \hline
    Rand ($\in\mathcal{D}_{\text{style}}$)    & {0.60} & {0.48} & {1.00} & 13.1 & 1.9 & 0.6 & 0.37 & {9.0} & {0.15} & {0.62} \\ 
    Retrieval ($\in\mathcal{D}_{\text{style}}$) & 0.20 & 0.21 & 0.10 & 10.7 & 1.7 & 0.7 & 0.45 & 6.5 & 0.04 & 0.15 \\
    Human ($\in\mathcal{D}_{\text{test}}$)  & 0.46 & 0.43 & 0.67 & 26.5 & 13.7 & 9.2 & 6.65 & 9.3 & 0.16 & 0.60 \\
            \hline
    S2S+LM & 0.50 & 0.44 & \textbf{0.59} & 16.3 & 3.0 & 0.8 & 0.44 & \textbf{8.7} & \textbf{0.07} & \textbf{0.38} \\
    MTask  & 0.17 & 0.22 & 0.14 & 19.5 & 4.5 & 1.5 & 0.73 & 6.9 & 0.03 & 0.12 \\
    \multicolumn{1}{r|}{+$\mathcal{L}_\text{conv}$ (\textsc{SpaceFusion})} & 0.39 & 0.33 & 0.22 & 18.9 & 4.6 & 1.5 & \textbf{0.76} & 7.7 & 0.04 & 0.17 \\ 
    \multicolumn{1}{r|}{+$\mathcal{L}_\text{style}$ (\textsc{StyleFusion})} & \textbf{0.55} & \textbf{0.48} & 0.47 & \textbf{20.6} & \textbf{5.1} & \textbf{1.6} & 0.73 & 7.8 & 0.04 & 0.17 \\ 
    \hline
    \end{tabular}
    \caption{Performance of each model on automatic measures. The highest score for generative models in each column is in \textbf{bold} for each target. the "count" metric is normalized by the value of the targeted style dataset. Note that the unit for BLEU is percentage. }
    \label{table:auto}
\end{table*}

\subsection{Ablation study and automatic evalution}

The automatic evaluation results for arXiv-like and Holmes-like response generation tasks are presented in Table~\ref{table:auto}. In both instances, \textsc{StyleFusion} achieved relatively high BLEU, and showed high style intensity. The Rand baseline has the highest style intensity but lowest relevance. 
S2S+LM has the comparable style intensity to \textsc{StyleFusion} but BLEU is much lower, consistent with the observation made by \cite{niu2018polite}. MTask shows significantly less style intensity than \textsc{StyleFusion}. Moreover, MTask's diversity, as measured by entropy4 and distinct1,2, is much lower, indicating that outputs of this model tend to be bland. Adding $\mathcal{L}_\text{conv}$ regularization, which is \textsc{SpaceFusion}, increases diversity, relevance and style intensity slightly, consistent with the finding in \cite{jointly2019gao}. Style intensity is further boosted by the addition of term $\mathcal{L}_\text{style}$. relevancy and diversity are not significantly affected by the addition of $\mathcal{L}_\text{style}$.

\section{Conclusion}

We propose a regularized multi-task learning approach, \textsc{StyleFusion}, that bridges conversation models and non-parallel style transfer by structuring a shared latent space. This structure allows the system to generate stylized relevant responses by sampling in the neighborhood of the model prediction, and to continuously control style intensity by modulating the sampling radius. We demonstrate this method in two tasks: generating arXiv-like and Holmes-like conversational responses. Automatic and human evaluation show that, without sacrificing relevance, the system generates responses of the targeted style and outperforms competitive baselines. In future work, we will use this technique to distill information from other non-parallel datasets, such as external informative text \cite{qin2019conversing, galley2019grounded}.

\bibliography{ref}
\bibliographystyle{acl_natbib}

\end{document}